\documentclass[runningheads]{llncs}

\usepackage[english]{babel}

\usepackage[letterpaper,top=2cm,bottom=2cm,left=3cm,right=3cm,marginparwidth=1.75cm]{geometry}

\usepackage{amsmath}
\usepackage{graphicx}
\usepackage[colorlinks=true, allcolors=blue]{hyperref}

\usepackage{academicons}
\usepackage{orcidlink}
\usepackage{todonotes}
\usepackage{float}
\usepackage[sort,nocompress]{cite}
\usepackage{graphicx}
\usepackage{subcaption}
\usepackage{xcolor}

\title{Towards the Visualization of Aggregated Class Activation Maps to Analyse the Global Contribution of Class Features}
\author{
Igor Cherepanov$^{1}$\orcidlink{0000-0001-9109-090X} \and
David Sessler$^{1}$\orcidlink{0009-0006-0258-7927}\and
Alex Ulmer$^{1}$\orcidlink{0009-0008-3778-8184}\and
Hendrik Lücke-Tieke$^{1}$\orcidlink{0000-0002-0934-6820}\and 
Jörn Kohlhammer$^{1,2}$\orcidlink{0000-0003-1706-8979} 
}
\institute{Fraunhofer IGD, 64283 Darmstadt, Germany\and
Technische Universität Darmstadt, 64289 Darmstadt, Germany\\
\email{\{igor.cherepanov, david.sessler, alex.ulmer, hendrik.luecke-tieke, joern.kohlhammer\}@igd.fraunhofer.de}} 
\date{
}
\authorrunning{I. Cherepanov et al.}
\titlerunning{Towards the Visualization of Aggregated Class Activation Maps}
\begin{document}
\maketitle

\begin{abstract}
Deep learning (DL) models achieve remarkable performance in classification tasks.
However, models with high complexity can not be used in many risk-sensitive applications unless a comprehensible explanation is presented.
Explainable artificial intelligence (xAI) focuses on the research to explain the decision-making of AI systems like DL.
We extend a recent method of Class Activation Maps (CAMs) which visualizes the importance of each feature of a data sample contributing to the classification.
In this paper, we aggregate CAMs from multiple samples to show a global explanation of the classification for semantically structured data.
The aggregation allows the analyst to make sophisticated assumptions and analyze them with further drill-down visualizations. 
Our visual representation for the global CAM illustrates the impact of each feature with a square glyph containing two indicators.
The color of the square indicates the classification impact of this feature.
The size of the filled square describes the variability of the impact between single samples.
For interesting features that require further analysis, a detailed view is necessary that provides the distribution of these values.
We propose an interactive histogram to filter samples and refine the CAM to show relevant samples only. 
Our approach allows an analyst to detect important features of high-dimensional data and derive adjustments to the AI model based on our global explanation visualization.

\end{abstract}
\section{Introduction}
Machine learning has progressed from simple algorithms to highly complex models.
While early approaches could be interpreted easily, new and more sophisticated models are difficult to explain.
The architecture of simple models, e.g., decision trees, is comprehensible and its decision-making process can be explained by the decision path in the tree.
More recent deep learning (DL) models based neural networks outperform simple models in many areas such as medicine, autonomous robots and vehicles, speech, audio and image processing.
These DL models are constructed by many nested and interconnected neurons that are able to identify complex patterns in data.
However, just the model performance is not sufficient for risk-sensitive applications.
Transparency and trust in the models are considered very important not only for ML developers but also for the domain experts who use these models.
Therefore, more attention is put into research on how to explain complex artificial intelligence (AI) approaches. 
Explainable artificial intelligence (xAI) focuses on techniques and algorithms to provide high-quality interpretable, human-understandable explanations of AI decisions.
This builds trust in the models, helps to better understand the models, and allows to evaluate the models by domain experts.
Based on this, a higher accuracy  and correctness of models is achievable.

The decision explanations can be described locally, providing an explanation for a particular classified sample, as well as globally, describing the importance of the features as a whole.
One of these techniques is called class activation map (CAM) and has its origin in image classification.
The CAM approach is restricted to convolutional neural networks (CNN) with a global average pooling (GAP) layer after the last convolutional layer and has the advantage of a local explanation that is directly calculated from the trained CNN model.
A CAM visualizes the impacting regions of an image that led to the specific classification by linearly combining activation maps (also called feature maps) on the last layer with the weights of the last fully connected layer corresponding to a target class neuron.

We transfer this approach to semantically structured data and extend it to aggregated CAMs with an interactive visualization approach.
Semantically structured data has a predefined structure where the order or position of each information unit is fixed. 
For example, the header information for network packets has a defined structure.
Because the data is semantically structured, an aggregation of CAMs is feasible and we can derive global information about the classification.
Our approach scales well with high-dimensional data and provides an overview visualization and an interactive histogram to filter relevant parts to improve the analysis.

Our visualization design is based on the established visualization mantra by Shneiderman~\cite{Shneiderman96}: "Overview first, zoom and filter, details on demand."
The goal is to provide an overview for a global explanation of the features that had a strong or weak impact on the classification for a particular class.
This overview is calculated on the basis of the aggregated CAMs.
Then, the analyst can select single features that are interesting and view the distribution of impacts for all samples for this feature in a histogram.
Finally, the histogram can be interactively used as a filter to exclude or select parts of the CAM samples and show a refined aggregation visualization.
The iterative approach of a drill-down can be leveraged to refine a CAM and the underlying patterns it reveals, enabling further examination and analysis.
Overall, our contributions are:
\begin{enumerate}
    \item An aggregation approach for CAMs as a global explanation technique for the classification of semantically structured data.
    \item A visualization design to show two key indicators for a fast overview of the global explainability of a class.
    \item An interactive approach to filter relevant samples to create a more detailed level of explainability. 
\end{enumerate}

\section{Related Work}

We group the work that is related. 
We first discuss xAI approaches, before we look at related work that proposes visual techniques.

\subsection{Explainable AI}
There is a wide range of different methods for xAI~\cite{angelov2021explainable,tjoa2020survey,molnar2022}. 
In this section, we will review the most common methods, that are also intended for a local explanation, and through the extensions can also be used for a global explanation.

One of the established methods is local interpretable model-agnostic explanations (LIME)~\cite{ribeiro2016should}.
This approach builds local surrogate models that are interpretable with the goal to approximate the individual predictions of the underlying complex model.
In this approach, new data points are created consisting of perturbed samples and the corresponding predictions of the black box model. 
These newly generated samples are weighted based on the closeness to the corresponding point.
Then LIME trains an interpretable model on this new dataset.
Based on this interpretable model, the prediction of the black box model is explained.
In this approach, an approximated model is built for the local explanations but it does not have to be a reliable global explanation.
Furthermore, LIME indicates the instability of the explanations~\cite{alvarez2018robustness}.

Another approach inspired by Shapley~\cite{shapley1951notes} applied in cooperative game theory and has been adapted for use in xAI to attribute the contribution of each feature to an individual prediction.
The goal of Shapley values is to estimate the contributions to the final model outcome from each feature separately among all possible feature combinations while preserving the sum of contributions being equal to the final outcome.
The calculation of Shapley values is only feasible for low-dimensional data.
For multi-dimensional data KernelSHAP and TreeSHAP were presented by Lundberg et al.~\cite{NIPS2017_8a20a862} to compute approximated Shapley values.
The KernelSHAP is a kernel-based estimation approach for Shapley values inspired by local surrogate models from the aforementioned proposed LIME approach~\cite{ribeiro2016should}. 
The TreeSHAP reduces the computational complexity but works only with tree-based ML models such as decision trees, random forests and gradient boosted trees.
Based on the work of Lundberg et al.~\cite{NIPS2017_8a20a862} some further modified KernelSHAP variants of this method were proposed~\cite{aas2021explaining} which considers feature dependencies in data.
These approaches provide impact values for each feature for an individual sample prediction.
The global impact is determined also by averaging the absolute Shapley values per feature in the data, similar to our approach.
Then the resulting values can be plotted in a bar chart, sorted by decreasing impact.
However, the variability among the absolute Shapley values is not considered. 
Furthermore, the Shapley values of all samples can be visualized in the so-called summary plot, which illustrates the distribution of the Shapley values per feature.
In this plot, the y-axis is determined by feature name and the Shapley values of each sample are located on the x-axis. 
These visualizations are used only to explain the model decisions for one class. 

Methods like LIME or SHAP are based on an explanation by learning an interpretable approximated model locally around the prediction. 
Our approach is based on a method that is extracted directly from the learned model which is applied to CNN models and is called Class Activation Map (CAM)~\cite{DBLP:journals/corr/ZhouKLOT15}.
CAM is an explanation technique used in computer vision to interpret and understand the decisions made by convolutional neural networks (CNNs) using global average pooling (GAP) for image classification tasks. 
CAM provides a saliency map of impacting regions of an image that contributed most to the prediction of a particular class.
The GAP is performed on the convolutional feature maps from the last layer.
These resulting values are used as features for a fully-connected layer that produces the desired output layer.
We obtain the resulting CAM by summing convolutional feature maps multiplied by the back-projected weights of the output layer.
The use of CAM was applied to the time series data~\cite{7966039} as well as to network data~\cite{pcapcam}.
For this type of data, it is meaningful to leave one dimension of the kernels at 1 since the input data samples represent a vector.
In work by Cherepanov et al.~\cite{pcapcam}, the CAMs were already aggregated by averaging to analyze the global differences between the network application classes.
In their work, network experts were interviewed. They said that a single CAM is simple and intuitive to understand and fits within the alignment of the hexadecimal representation of PCAPs.
In this work, we extend the visualization of aggregation and also provide several methods for an analyst to build the resulting CAM for a global explanation.

\subsection{Visualization}

Information visualization plays a crucial role in assisting, controlling, refining, and comparing for various domain and ML experts at different stages of an ML pipeline~\cite{SACHA2017164, lu2017recent, 10.1145/2851581.2856492, doi:10.1177/1473871620904671, https://doi.org/10.1111/cgf.13092}. 
From the beginning of this pipeline, these visualization approaches can be used to facilitate the understanding of complex raw data.
Visualization of data~\cite{8405549}, and data transformation~\cite{8851280}, allow experts to gain insight, identify patterns, and uncover relationships~\cite{7536217}.
For selecting a fitting algorithm, visualization methods provide a valuable means of comparison. 
Experts can interactively benchmark the performance of different algorithms, allowing them to make informed decisions based on their specific needs and goals~\cite{8781593,https://doi.org/10.1111/cgf.14301}. 
In this way, experts can select the most appropriate algorithm for their tasks, promoting efficient and effective analysis.
The information visualization and visual analytics research fields propose a number of approaches that cover these points~\cite{https://doi.org/10.1111/cgf.14329, https://doi.org/10.1111/cgf.14034}.
Furthermore, visualization facilitates the fine-tuning of parameters for various methods. 
By visually evaluating the impact of different parameter settings on the results, ML experts can optimize the performance of their chosen algorithms~\cite{https://doi.org/10.1111/cgf.14300}. 
This iterative process empowers users to refine and enhance their models, leading to improved accuracy and reliability.
There are also some approaches that visually explain how some ML methods work~\cite{8440091, vig-2019-multiscale, YU2018147}.
Finally, visualization serves as a means to represent the results of analyses and predictions made by models. It enables domain experts to communicate and comprehend the outcomes and diagnose the model and identify
problems effectively as proposed in work by Collaris et al.~\cite{9086281}. 
Additionally, the integration of xAI techniques into visualization empowers domain analysts to understand the predictions made by models in a transparent and interpretable manner~\cite{pcapcam, 10.1145/3301275.3302265, 8807299}. 
This ensures that users can trust and comprehend the decision-making process of the model, enhancing their confidence in the results~\cite{https://doi.org/10.1111/cgf.14034}.

In the past heatmaps were mostly used for CAM visualization, hence, we highlight related work from the visualization and machine learning communities.
First, we show visualizations for CAMs and then compare them to high-dimensional heatmap visualizations for other application domains.
With that, we explain the advantages and shortcomings of these approaches and provide a design rationale based on the findings.

CAMs and their visualization originated from the image classification domain, where the goal is to show which pixels lead to the detection of an object in an image.
Therefore, the image is overlayed with a semi-transparent heatmap using a rainbow color map~\cite{DBLP:journals/corr/ZhouKLOT15}.
Orange and red colors show a high recognition rate while purple and blue show no signal. 
Recent approaches show how these visualizations are used extensively to explain and improve the recognition models~\cite{jiang2021layercam,muhammad2020eigen,yang2019towards}.
While this visualization method has its benefits, all of the approaches use a rainbow color map.
Research in the visualization and cognition community showed that this color map has a lack of perceptual ordering, often misleading the interpretation~\cite{borland2007rainbow}.
The data domain for CAMs can be normalized to a range around zero, e.g. -1 to +1, where -1 is an indicator that this feature does not represent a class and +1 is a clear indicator for a class.
Research has shown that for this diverging data domain, certain colors are more suitable for human perception~\cite{harrower2003colorbrewer}.
One of these is for example the blue to red color map with a neutral white in the middle.
Based on this we adapted our visualization of CAMs. 
This color map is also often used in high-dimensional heatmap visualization in other research domains~\cite{blumenschein2018smartexplore}.
For structured data, multiple approaches have shown a matrix-styled heatmap, highlighting the separation of single features~\cite{clustvis}. 
For such high-dimensional visualizations interactive features like selecting, sorting and filtering are very important~\cite{fernandez2017clustergrammer}.
Other works from the visualization community extended the matrix heatmap visualization with additional visual indicators besides color~\cite{RuppertBULK14, blumenschein2018smartexplore } or extended it to hexagonal maps~\cite{trautner2022honeycomb}.
Based on these approaches we used the size of the filled rectangle in a matrix cell to visualize another quantitative value.
This improves the fast overview of the data and is beneficial for our use case of aggregated global CAMs as we can also show the variability of the aggregation. 

Finally, the interactivity of such an aggregated visualization is important~\cite{Shneiderman96}. 
The user has to be able to focus the analysis on relevant parts of the data and see the raw data to understand the aggregation.
We designed the visual-interactive approach based on the established visualization mantra.

Based on the related work we extend the CAM visualization and add new interactive elements to improve the analysis workflow.
According to the best of our knowledge, none of the recent approaches aggregate CAMs and visualize them the way we do in this paper.

\section{Approach}
\subsection{Semantically Structured Data}
Semantically structured data is a special form of structured data where the order or position of each information unit is fixed. 
Consequently, it becomes feasible to aggregate local explanations in the form of CAMs for each information unit or feature across multiple samples.
One example of this is network traffic data that has a specific protocol header information with a fixed order of units.
Images on the contrary are not semantically structured because an object can be at different parts of the image. 
A difference to structured high-dimensional data is that semantically structured data should not be resorted for the visualization because neighborhood information is lost.
Our approach is applicable if the aggregated features with the same semantic have the same position among all samples.

\begin{figure}[htb!]
\centering
 \includegraphics[width=1\linewidth]{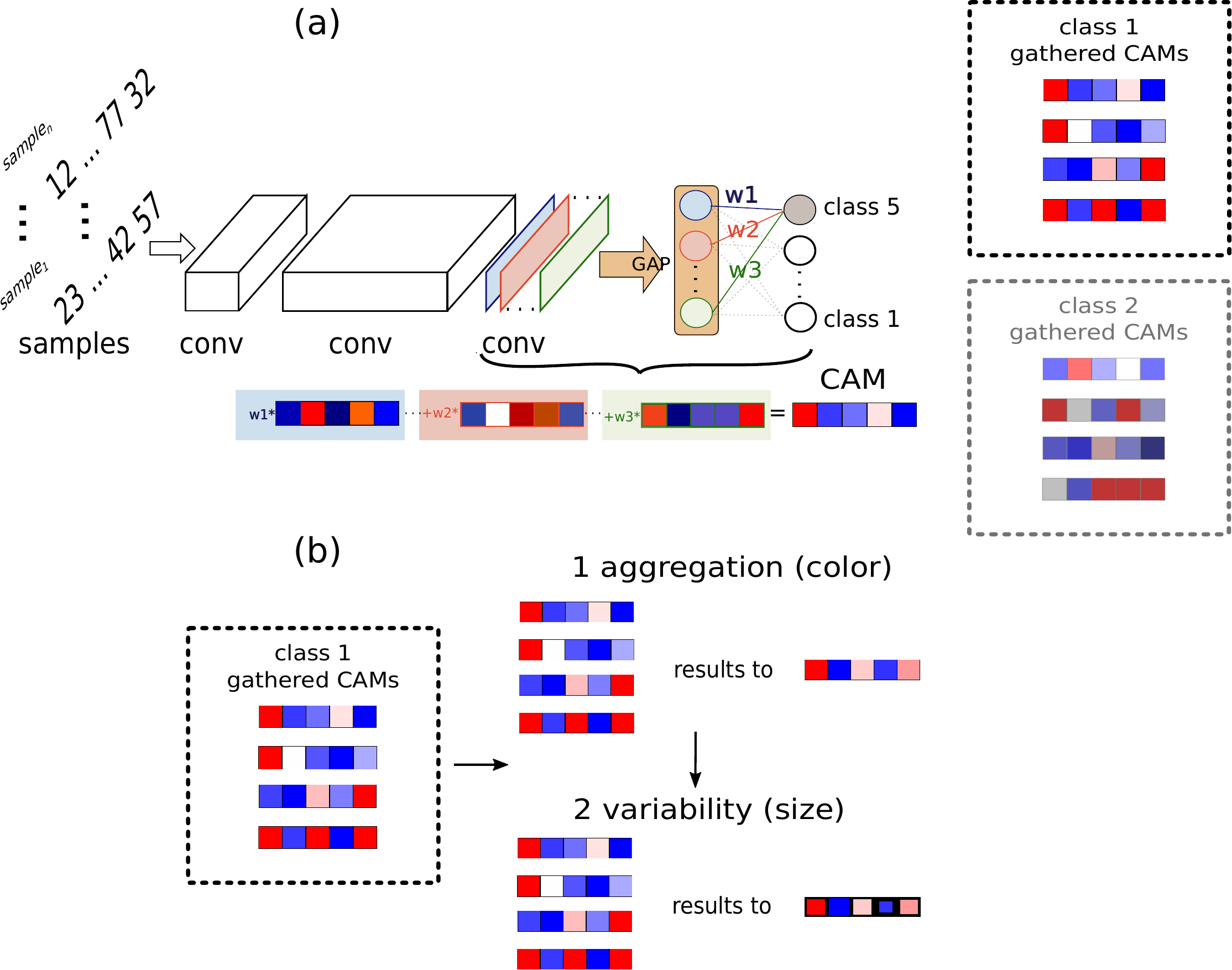}
\caption{(\textbf{a}) represents the calculation of a CAM for \textbf{the local explanation of a sample}. A CAM is obtained by taking the output of the last convolutional layer of the CNN and applying global average pooling (GAP) to reduce the spatial dimensions of the feature map to a single value per channel. This results in a feature vector that represents the importance of each channel in the final prediction. The feature vector is then passed through a softmax activation to obtain the class probabilities. The weighted sum of the feature maps, where the weights are the class probabilities, is then computed to obtain the final heatmap. (\textbf{b}) represents our approach to \textbf{aggregating the CAMs for a specific class}. 
The result of the aggregation is represented by two indicators. 
The first one aggregates the impact values for a feature, which is represented by the coloring. For the second the variability is calculated, which is represented by the size of a grid. 
}
\label{camandagg}
\end{figure}

\subsection{Technical Background} \label{technicalback}

The CAM technique originated from image classification in the field of computer vision.
In the case of images, each pixel is considered as input.
The goal of the method is to highlight regions of an image sample that had the highest impact on the classification of the predicted class.
The CAMs are extracted from a CNN model.
Our approach focuses on semantically structured data in general.
Therefore, we process each unit of transformed data as an input to the model.
For example, if a sample is represented in bytes, each byte is considered as input.
The kernels are not suitable with the second dimension greater than 1.
Because the input sample is considered as an 1D vector and not as an image, where the upper and lower neighboring pixels can be related to each other. 
In the case of an 1D vector, only the neighboring properties in the same axis might be related to each other.
For this reason, 1D kernels are applied with different lengths for CNN in our approach.
Global average pooling (GAP) is performed on the feature maps from the last convolutional layer.
Then, the resulting values after GAP are connected to the final fully-connected layer that produces the desired classification output.
This structure allows us to construct a CAM by projecting back the weights of the output layer on the convolutional feature maps.
In this way, we calculate the resulting CAM as a sum of all convolutional feature maps of the last convolutional layer multiplied by the weights of the output layer. 
The calculation of CAM is illustrated in Figure~\ref{camandagg}a.

\subsection{Aggregation of CAMs}
The next step is to aggregate the CAMs of each predicted class. 
A number of CAMs of the same class are collected into an array.
The result is a 2D array in which each row represents a CAM (horizontal axis) and each column represents the impact of the feature in a CAM (vertical axis).
Our next task is to build an aggregated CAM that represents all these CAMs considering the values of the vertical axis as well as the variability among them. 
This procedure is shown in Figure~\ref{camandagg}b.

We aggregate the values along the vertical axis to describe the importance of each feature with a single value, building the global CAM in the process.
For this, we provide multiple methods for the aggregation.
The result of these methods can represent different aspects of the impact distribution in the classification analysis.
One of these methods is the calculation of the mean which can be used to assess the overall average of the distribution.
An alternative method is the median of the CAM values.
This serves a similar purpose but is less prone to be influenced by extreme values, that might be outliers.
When two or more peaks in the density of the impact distribution exist, methods for calculating the global mode of a density are meaningful.
For this, we apply the kernel density estimation method and take the most frequent value of the density~\cite{10.1214/21-EJS1972}.

In a potential application, it is reasonable to provide all these aggregation methods since they enable different types of analysis for users.
Additionally, we calculate a second indicator that represents the variability of the feature impacts between all CAMs for the individual class.
This value is used to assess if the range of the impact values is significantly predominant between all local CAMs of the considered class or the opposite, the values are highly spread out.
As the indicator of variability, we implemented and tested variance, standard deviation, entropy, and the Gini coefficient.
Before the calculation is applied, we normalize the values to [0,1] to achieve comparability between the resulting variability values.
After the calculation, in the case of variance and standard deviation, the results are normalized again to [0,1] because the resulting value can be in the range of 0 to 0.25 or 0.5 respectively.
This is done to ensure the same value domains for all variability indicators.
After testing these four variability measures, we found that the values calculated through entropy spanned the value range more evenly and thus were best suited for our dataset.
However, all these variability measures describe different characteristics of variability and therefore should all be included as a user parameter in a potential application.

\subsection{Visualization}
Getting a better global understanding of CNN model decisions in classification tasks is the main goal of this work.
Visualizing numerical and statistical data has proven to enhance the capability of humans to understand complex subjects. 
This is the reason why visualizing the impact values and their distribution is an essential part of our approach.
We utilize visualizations to provide an overview over large amounts of data points while simultaneously allowing the user to detect interesting patterns at a glance.   
\begin{figure}[htb!]
\centering
  \includegraphics[width=.9\linewidth]{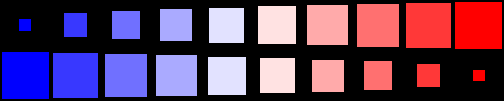}
  \caption{Overview of visual mapping of the two indicators. The color represents the aggregated feature impact values while the size illustrates the variability of the impact distribution.}
  \label{fig_colormap}
\end{figure}
For the visualization of distributions box plots or violin plots are commonly used~\cite{hintze1998violin}.
However, while these plots work great to display and compare multiple distributions of values, they only work for the comparison of relatively few distributions.
Since for our use case, a comparison of a hundred or more distributions is necessary we decided to propose a different solution.
To make the visualization scale better for our needs we map the distributions to two indicators, it is aggregation and variability, before visualizing them.
Further, we suggest interactive visualizations that reveal additional information about underlying data and its distributions on demand.
This allows the analysts to drill-down and explore interesting regions of our explainability approach to gain a more robust understanding of the model behavior.
In detail, we start with a visualization of the global aggregated CAMs including their variability values.
For this, we transform the long one-dimensional vector of aggregated impact values into lines that we align on top of each other. 
This results in a structure similar to a text passage in a book where the features are ordered from left to right in a line and the vertical orientation of the lines is from top to bottom. 
Each grid cell in this visualization contains a square patch that represents two values, the aggregation and the variability, which we presented in the previous Section~\ref{technicalback}.
We map the aggregated impact value for the feature importance to color.
This way, it enables the analysts to efficiently detect patterns or regions in the long vector that are either contributing the most to the prediction of a class or are notable for other reasons.
We choose a divergent color map as seen in Figure~\ref{fig_colormap} which maps features with the highest aggregated values to the color red, the features with the smallest aggregated value to blue, and the ones in the center of the value range to white.
This coloring choice clearly indicates the most and least impactful features for the classification while also highlighting features where the model provides mixed results.
We also considered classical color maps often used in the machine learning community like jet, viridis and turbo~\cite{borland2007rainbow,liu2018somewhere,reda2020rainbows}.
We discarded the former because of its issues regarding the brightness profile and the unsuitability for color-blind people.
The latter two, while providing a higher distinguishability between values due to the use of a wider range of colors, could not satisfy our need to highlight the values directly in the center of the value range.
Therefore, our choice was to implement a diverging color map, that keeps the colors familiar in the machine learning community at the start and the end of the spectrum.
This design choice was also motivated by the work of Moreland~\cite{moreland2009diverging} which reflects on the construction of diverging color maps.
The white color in the center of the spectrum represents distributions where blue and red values in the features CAMs have a similar presence or mutually aggregate to a neutral value.
Investigating the impact distributions of such mixed cells might yield insights into the prediction mechanisms of the model.
The second parameter, the impact variability value, is mapped on the size of the square patch which is centered in each grid cell. 
An alternative visual representation that we considered for the representation of the impact variability was a bar plot.
It supports the comparison of cells that are one-dimensionally orientated.
However, we discarded this option because for the alignment of cells in 2D the benefits of a bar plot versus a square representation are mitigated.
Further, the bar charts, starting at the bottom of each cell, created visual disorder, especially for the representation of a large amount of aggregated impact values.
The centering of the square patches avoids this negative effect (see Figure~\ref{fig_colormap}).
It is important to define a minimum size for the squares, to still see the color of the aggregation values represented.
This way analysts can inspect the impact values of the feature regardless of their variability. 
Further, we choose the size of the squares based on their area to accommodate the quadratic interdependency of the side length to area of the squares. 
This leads to a perceptually intuitive mapping of the numeric values to the visual representation. 
To display the actual values of the aggregation and variability, a tooltip should be provided, that can be accessed by hovering over one of the grid cells.\\
\begin{figure}[htb!]
\centering
  \includegraphics[width=.8\linewidth]{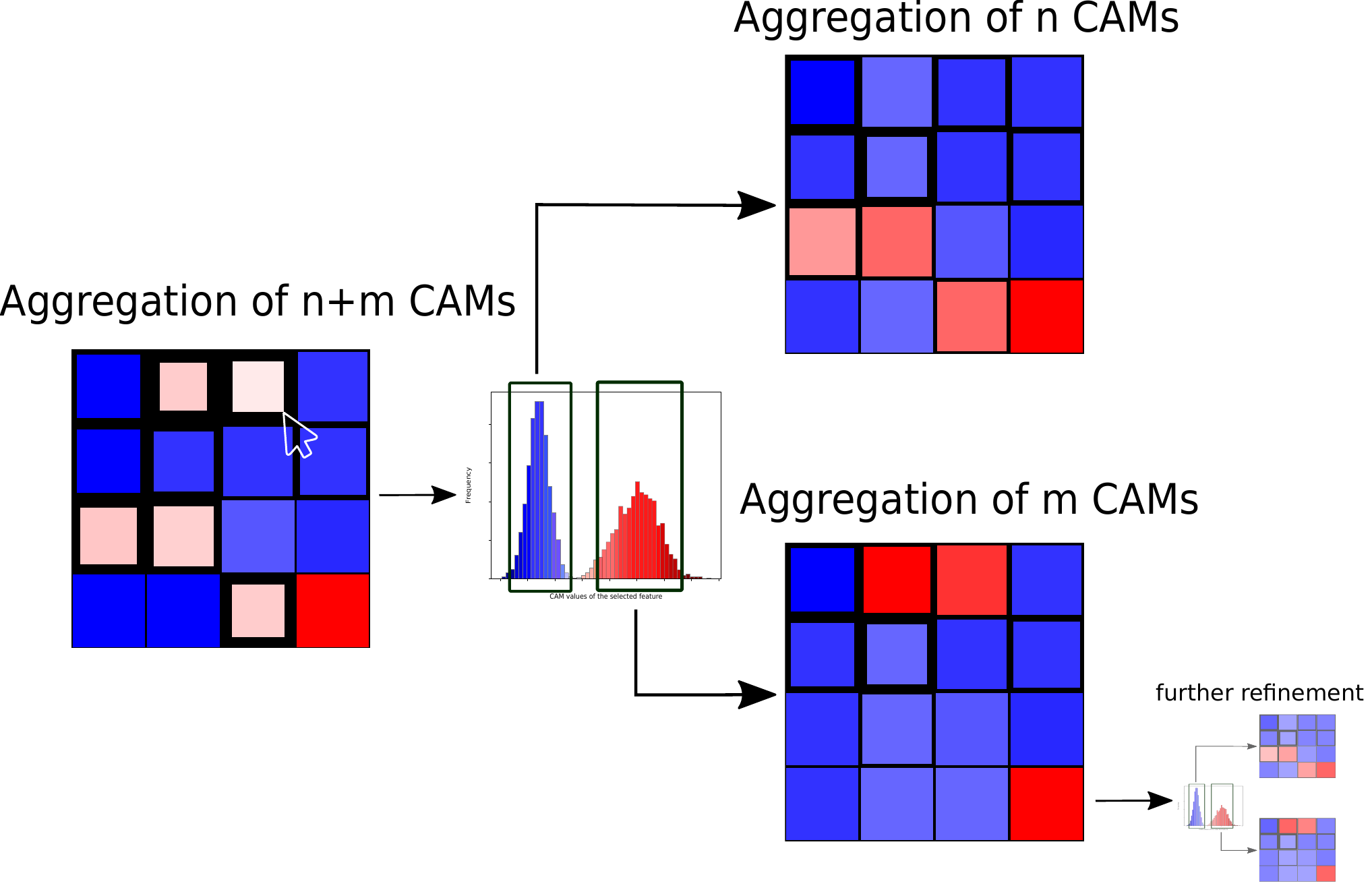}
  \caption{\textbf{Visual-interactive drill-down:} Clicking on a colored cell in the grid shows the histogram of impact values in this cell. A range selection control allows filtering of the samples that then are used to display a CAM based on the aggregation of the selected subset. This interaction can be iteratively repeated.}
  \label{fig_drill_down}
\end{figure}
Users can interactively select the aggregation and variability measures.
This way, they can configure the visualization outputs to their analysis needs.
To investigate distributions of interesting impact value aggregations the analysts can click on the corresponding cell.
This interaction shows a histogram plot of the aggregated impact values as depicted in Figure~\ref{fig_drill_down}.
Here, the analysts can explore the distribution characteristics and interactively select regions of the distribution with a range selector control.
This interaction can be meaningful for features, where the histogram indicates multiple modes in the impact value distribution.
Multiple modes suggest, that the same feature entails different impact indicators which is an interesting case worth analyzing.
Selecting a range of the distribution provides a view of a sub-global CAM that has identical visual properties as the initially shown global CAM.
This sub-global CAM is aggregated based on the samples that correspond to the selection in the histogram as shown in Figure~\ref{fig_drill_down}.
Viewing such sub-global CAMs can help the analyst to understand if certain features in this class are not relevant for all samples or on the contrary relevant.
The process of drilling down the aggregated CAMs can be iteratively repeated until the displayed CAM is only based on a single sample.
Finally, interactive annotations of the inspected cells allow the analysts to mark cells during their exploration of the CAMs.
This way, interesting cells can be highlighted, while cells that did not contribute to the understanding of the model could be discarded by marking them as not relevant for the prediction.

\section{Usage Scenario}

In this usage scenario, we describe how a network domain expert can exploit our proposed approach of aggregated CAMs for global explanations of predicted application classes by a CNN model. 
The classification of network data has received wide attention both in the scientific community and in the industry.
Classifying network data is important for network security, network monitoring and management, traffic analysis, resource optimization, and cost management. 
Accurate classification of network data provides insights into the nature of network traffic, enabling organizations to ensure network security, optimize network performance, and effectively manage network resources.
For this reason, there is a wide variety of approaches for network traffic classification, including solutions based on deep learning~\cite{networkclassificationsurvey1,networkclassificationsurvey2,networkclassificationsurvey3,networkclassificationsurvey4}.
Since certain decisions based on these traffic classifications can be made, an explanation of the predictions is necessary.
xAI enables transparency, interpretability, fairness, accountability, and security of AI systems, as well as increases trust from network experts~\cite{burkart2021survey,das2020opportunities,gunning2019xai}.
Furthermore, through the explanations network experts can verify the correctness and robustness of the model.
By incorporating their feedback ML experts can improve the model.

In our scenario, ML experts provide a model for network analysts.
The model is based on a CNN architecture that performs best for network traffic data~\cite{aceto2019mobile,lotfollahi2020deep}.
The applied model has a similar structure as described in previous related work~\cite{pcapcam,lotfollahi2020deep}. 
The CNN consists of the following feature map dimensions in the hidden layers: 16, 32, 64, 128, 128, 128 with stride size of 1 and 1D kernel size of 5, followed by GAP and a fully connected layer with the same number of outputs as application classes in the dataset. 
The CNN model was trained using the categorical cross entropy as loss function and Adam optimizer.
The model is trained on the ISCX VPN-nonVPN dataset which is a widely used dataset in network classification research~\cite{wang2017end, iliyasu2019semi, datasetispx}. 
This dataset suffers from severe class imbalance, with the FTPS class having a significantly higher number of samples (7872K) compared to the AIM class (only 5K samples). 
To address this issue, we apply random undersampling to balance the classes~\cite{9200087}, aiming to have approximately 5K samples per class, following a similar approach used in the work of Lotfollahi et al.~\cite{lotfollahi2020deep,pcapcam}.
In the data preprocessing phase, the Ethernet header is removed, source and destination IP addresses are masked to 0.0.0.0 and the shorter samples are padded to a fixed size of 1500 bytes~\cite{lotfollahi2020deep,pcapcam}.
We train the model until it achieves solid performance - on the training dataset: F1: 93.4 Recall: 93.5, Precision: 93.7, and on the test dataset F1: 92.8 Recall: 92.8, Precision: 93.0.

\textbf{Pattern observation.} 
With the integration of the classification and its explanations in tools similar to Wireshark~\cite{ulmer2019netcapvis,chappell2010wireshark}, the experts are able to see the impact of each byte for the prediction of an application class.
A single CAM illustrates a local explanation of a classification.
This explanation is useful for a closer examination of individual network packets. 
While local explanations provide a finer-grained understanding of the model predictions on individual samples, global explanations generalize to a broader overview of the model behavior.
With our approach, we aim to explain a class globally by aggregating the CAMs for a particular class.
The global representation allows the analyst to see the differences between the classes and deduce insight from them.
The first indicator in our visualization is the color, which tells how many bytes in a PCAP contribute to a class prediction.
The second indicator is the size of the square, that reflects the variability of the impact values in the distribution. 
The average of impact values for the first indicator and for the second entropy are the most representative on the dataset.

\begin{figure}
            \centering
            \includegraphics[width=.8\textwidth]{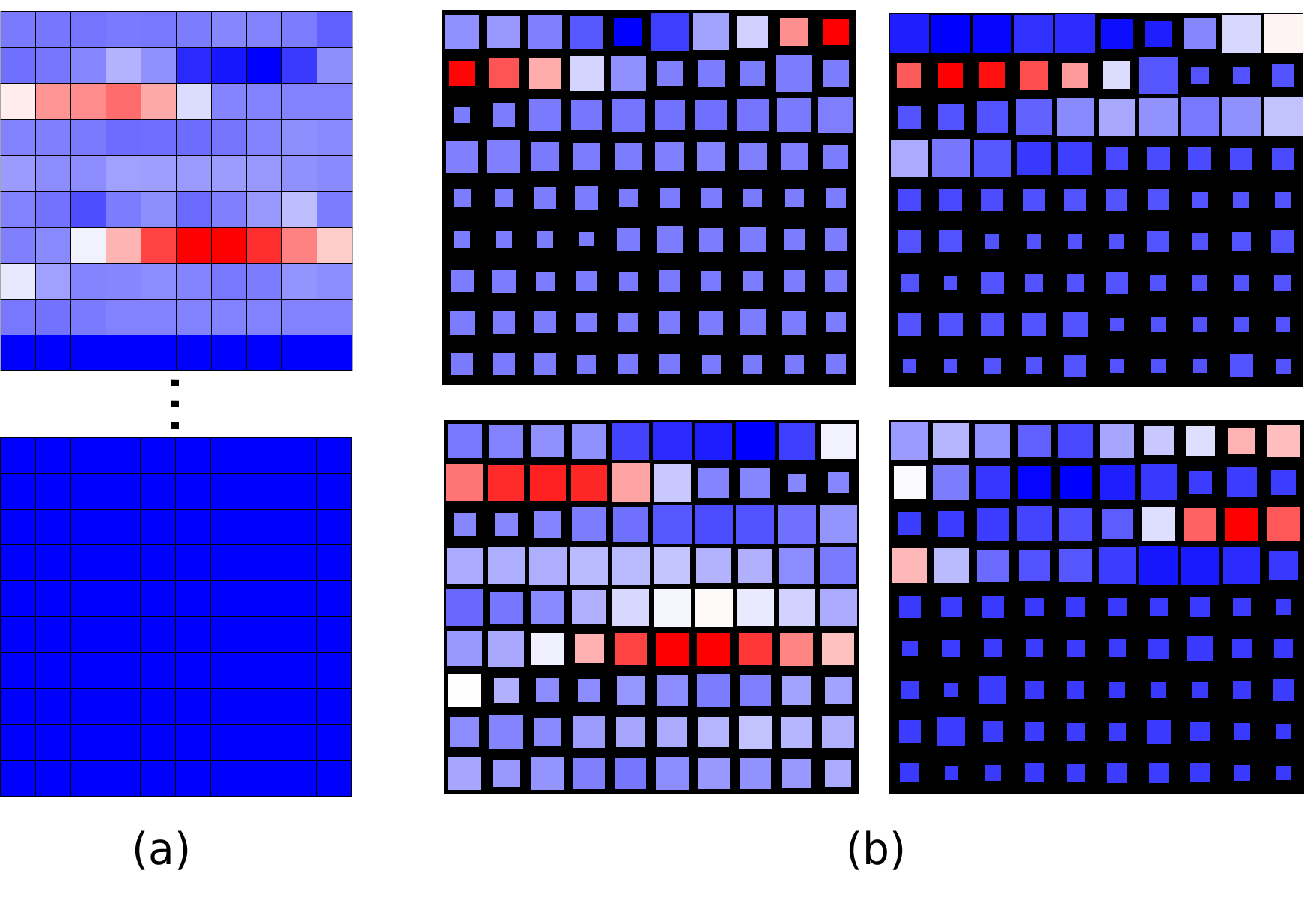}
    \caption{(\textbf{a}) A single CAM (10x150 dimensional) that represents a classified PCAP sample, the encrypted part of the CAM is cut out due to the size and irrelevance on classification.
    (\textbf{b}) Four aggregated CAMs represent distinct application classes of PCAPs. The aggregated CAMs are shortened to a smaller dimension since the encrypted part of PCAPs in the CAMs is always irrelevant for the predictions.}
\label{usagescenariopcap}
\end{figure}

The resulting global CAM with our provided visualization allows analysts to easily detect the impact of each feature.
The analysis of large red cells is the first step in the global CAM examination, as the analyst first confirms if the system is in line with his expert knowledge about network packets.
The patterns represent the significant bytes of the packets of this class.
The analyst examines distinct CAMs and starts to identify patterns among the application classes.
She observes that certain bytes or byte ranges have different patterns of high-importance features (red cells).
In Figure~\ref{usagescenariopcap}b four patterns of four distinct classes are represented. 
Some application classes share similar patterns of byte importance, indicating potential similarities in their network traffic behaviors, such as the 10-12th bytes in three classes except the last one (Figure~\ref{usagescenariopcap}b).
These refer to the protocol and header checksum in the IP header.
The first information can be quickly verified because specific applications use UDP, while others use TCP protocol.
Every verification that an expert can confirm through his knowledge strengthens his confidence in the AI system.
It is apparent that also other important bytes in the UDP or TCP header are significant for the classification. 
Next, the analyst examines the involved bytes in the header which represent the port number.
Because they are also significant to the application class the analyst can confirm that the system is correct in this case, which again raises the trust in the system.
Now more questionable impact values are analyzed.
Certain bytes located at the beginning of payloads are significant for classification as seen in Figure~\ref{usagescenariopcap}b in the third CAM visualization.
This indicates information presenting the header of the corresponding application which is investigated further, by using the distribution histogram to reduce the data to relevant parts.
The resulting patterns are analyzed by the expert and provide a comprehensive understanding of the data as well as the behavior of the model.
Important features and consequently their data content can be identified and reviewed and specific decisions can be deployed for the network based on the findings. 

\begin{figure}[H]
\centering
  \includegraphics[width=.8\linewidth]{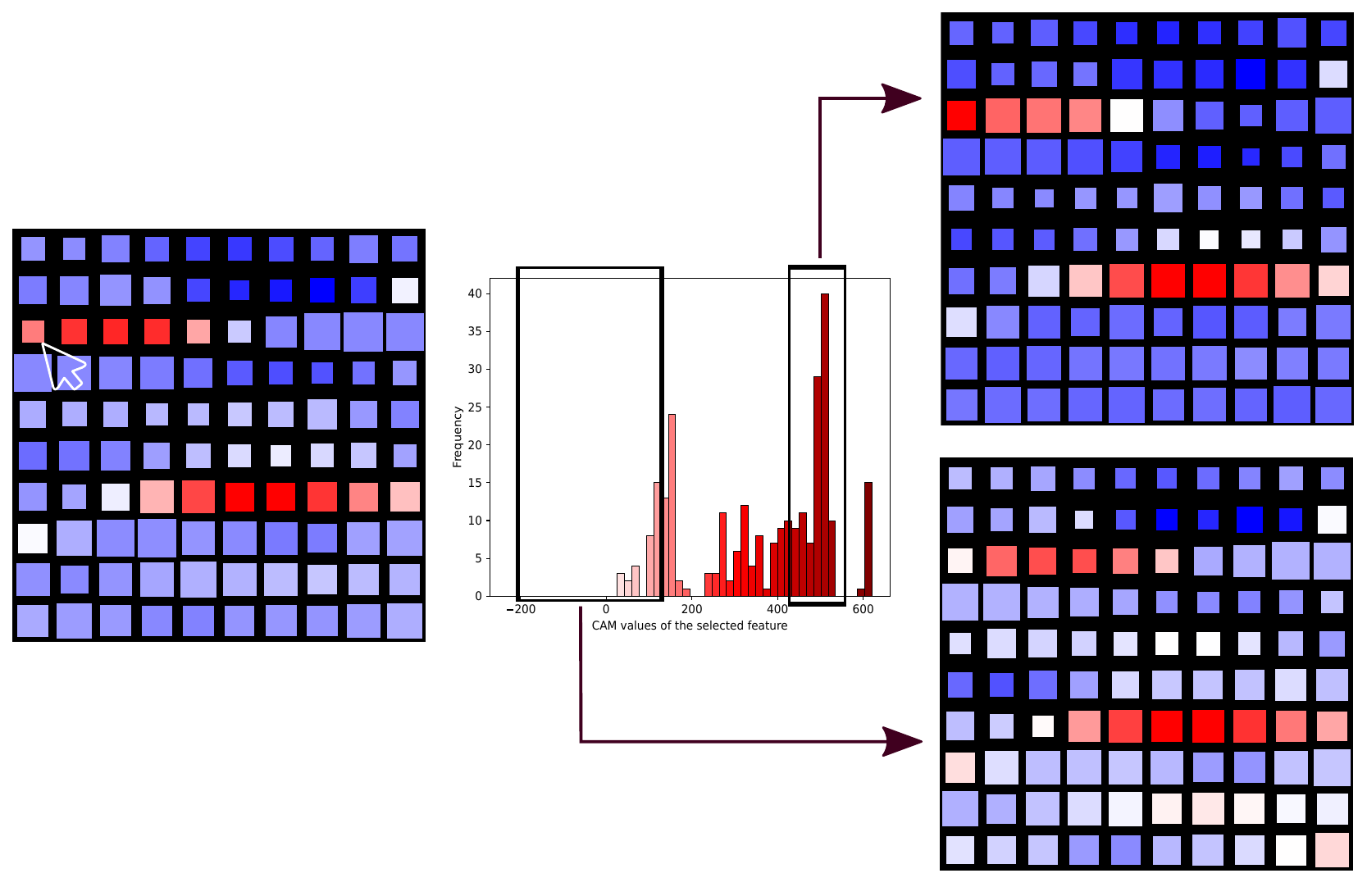}
  \caption{An expert refines the global CAM for an application class (ICQ) by examining the distribution of impact values by eliminating particular CAMs.}
  \label{drilldownusagescenario}
\end{figure}

\textbf{Drill-down Exploration.}
For multi-peak impact value distributions, it is useful to refine the aggregated CAMs. 
Specific impact distributions within a class can potentially cancel out each other through aggregation.
Cells that might contain such interference can be identified by the size of the squares.
This is because smaller squares indicate a strong variability of impact values between the local CAMs.
For example, if a feature of a class is significant in many samples, but, in others, this feature is not relevant for the classification.
For instance, this can be information in the header that is not present in all network packages of a class.
In this case, the network expert refines a global CAM for a class interactively based on a cell of interest with a smaller square.
By selecting this cell it is useful to see the distribution of CAM impact values (see Figure~\ref{drilldownusagescenario}).
If the visualization of the CAM values shows multiple modes in the distribution, it is meaningful to separate them from each other.
The expert can filter the histogram by selecting a unique mode. 
The selection of the unique mode provides a uniform distribution of the impact values. 
In this way, the local explanations are selected where this feature was impacting.
This allows the expert to filter out these packets with different impact features and investigate and compare their data content. 

\textbf{Model Improvement.} 
An individual CAM and aggregated CAMs are valuable in different contexts, and a combination of both provides a more comprehensive understanding of the model's behavior, strengths, and limitations.
Using the local CAM as well as the global aggregated CAMs allows analysts to check the expectations of classifications through their expertise. 
In the process of using the model, its correctness and performance can be verified.
In this way, the weaknesses of the classifications can be identified.
There are potential weaknesses, such as the difficulty to distinguish certain classes from each other.
Particular features of the input data could be misleading, such as the IP address, which is often eliminated from the input so as not to bias it~\cite{lotfollahi2020deep,pcapcam}. 
This avoids overfitting, moreover, IP addresses have a high volatility, which would invalidate the model immediately.
Furthermore, in the case of encrypted files, it is apparent that including the full 1500 bytes in the classification is not useful since the CAMs show that the encrypted part of the PCAPs is irrelevant for the classification of all PCAPs (see Figure~\ref{usagescenariopcap}a).
Thus, the analyst can keep the model much smaller in terms of the input data and consequently in terms of the total number of model parameters.
The constant innovations and changes of the applications can also cause a change in the data characteristics of network traffic, called data drift~\cite{ackerman2021automatically}.
When data drift occurs, the model performance may become less accurate or reliable.
The explainability that clarifies classifications to an analyst helps to diagnose the reasons for model performance degradation and provides insight into the factors that contribute to data drift.
In summary, aggregated CAMs help to diagnose, explain, and evaluate machine learning models, thereby enabling ML experts together with network analysts to maintain the reliability, correctness, and trustworthiness of a model.

\section{Discussion and Future Work}
Local CAMs provide an explanation of classification for one sample. 
With our approach, it is possible to aggregate CAMs  generalizing a global explainability for each class. 
Our proposed visualization for the resulting CAM, the significant values are visualized taking into account also the variance among the CAM values for each feature. 
Our visualization allows a user to quickly discover impacting features in high-dimensional data.
The resulting CAMs for each class represent the impacting values. 
Thus, it is possible to detect the most important features of a class and also to compare the patterns of these classes.
However, our approach is limited to the CNN models trained on the semantically structured data, since the position of the features plays a significant role for the aggregation.

A future research direction might be to compare the CAMs with a distance measure.
A resulting CAM is a two-dimensional vector, so by the distance measures also the similarities among the CAMs could be calculated.
An investigation into which metric would be suitable for this is also interesting. 
The vectors of a CAM could also be weighted according to the importance of the impacting value or the variance between the values.

\begin{figure}[ht]
\centering
\includegraphics[width=.31\textwidth]{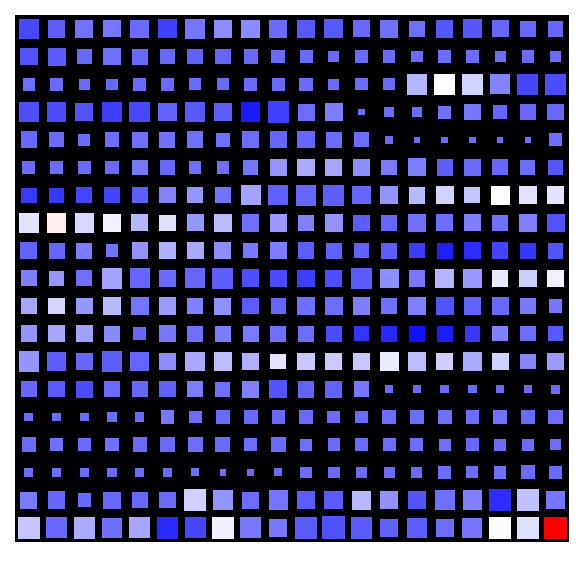}\quad
\includegraphics[width=.31\textwidth]{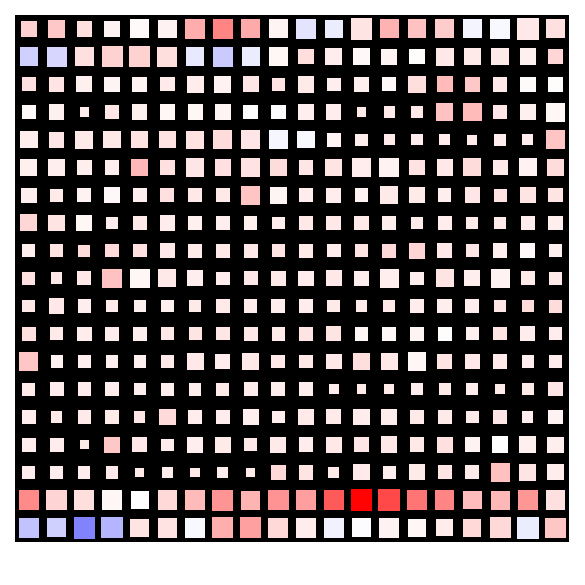}\quad
\includegraphics[width=.31\textwidth]{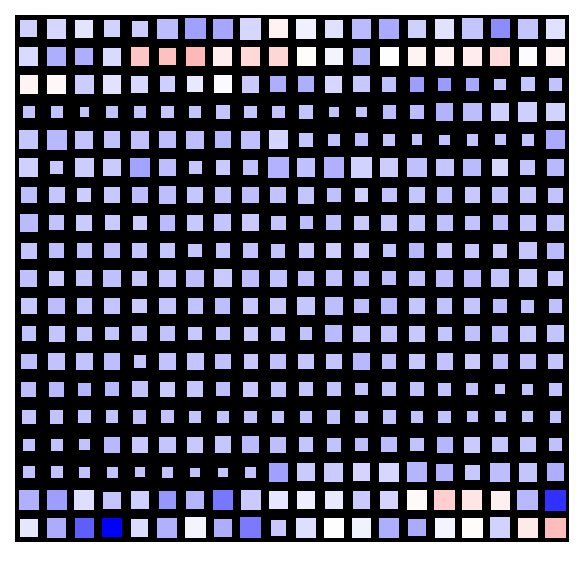}
\caption{Aggregated CAMs for three distinct classes of the UJIIndoorLoc dataset~\cite{torres2014ujiindoorloc}.}
\label{UJIIndoorLocpic}
\end{figure}

The patterns in a CAM may not always be easy to understand, which may prevent the extraction of simple patterns for each class.
For example, Figure~\ref{UJIIndoorLocpic} illustrates three classes from the  UJIIndoorLoc dataset~\cite{torres2014ujiindoorloc}, in which the classes are distinguishable in terms of impacting features, however, the patterns appear to be too complex. 
The patterns have many impact features that are spread in an unstructured way in the aggregated CAM.
Such patterns can possibly consist of different interferences in the aggregation caused by multiple modes in the impact distributions.
Further, CAM refinement is suitable for this.
The illustrated impact features in the global CAM reduce the possible starting points of selection for a CAM refinement.
However, this remains as not a simple problem and does not guarantee the best result after a user's modification.
It can also happen that the impacting properties are the same in distinct classes because they only show the position in the data and the content in this position must be examined by the user.

It would also be possible to sort the properties according to the feature impact value for further inspection if the data allows such a reordering. 
That way, correlating features would appear at similar locations, enhancing mental map building within the user, and therefore potentially ease the sense-making process.
However, not all data can be sorted in this way.
For network data such as PCAPs, each byte is an input into the CNN model and particular segments consist of multiple bytes~\cite{mandl2018tcp}. 

There is also an option to use our visualization approach to illustrate the correlation between the impact values of gathered local CAMs. 
The collection of CAMs represents a matrix (number of impact values x number of CAMs), this collection can also be used to construct a correlation matrix that represents the correlation between the impact values.
The correlation matrix would have the size number of impact values x number of impact values.
In this way, the correlations between the impact values in the CAMs can be obtained.
The correlations matrix can also be visualized with our proposed visualization where the colors represent the positive or negative correlation value and the size represents the absolute correlation value.
The correlation may possibly indicate which features have a synergetic effect on the model behavior.

In our ongoing research, we plan to further advance our proposed approach of aggregated CAMs by integrating it into a practical application that caters to the specific needs of domain experts.  
NetCapVis~\cite{pcapcam, ulmer2019netcapvis}, an application from our previous work is especially suitable for this.
There, we already evaluated the visualization of single CAMs for PCAP data with network experts.
The results were overwhelmingly positive.
We plan to integrate and evaluate our proposed approach with this application to confirm if experts can benefit from the global CAMs and their interactive refinement, allowing the classification to be interpreted and explained.
We will seek their perspectives on the usefulness of the explanations in understanding network traffic patterns.
Specific aspects that will be evaluated include: 
\begin{itemize}
    \item \textbf{Interpretability}: the extent to which the global explanations are understandable and interpretable by network experts, allowing them to gain insights into the underlying factors influencing the classification of different application classes.
    \item \textbf{Relevance}: the relevance of the explanations to the domain knowledge and expertise of the network experts, ensuring that the explanations align with their expectations and provide valuable insights specific to their field.
    \item \textbf{Actionability}: the practical utility of the explanations in enabling the network experts to take proactive actions, such as implementing targeted security measures, refining network configurations, or detecting and mitigating potential security threats effectively. This enables to validate and improve the resulting model by ML experts.
    \item \textbf{Usability}: including its interactivity, we aim to ensure that it provides a seamless and user-friendly experience for network experts.
\end{itemize}   
We also plan to conduct evaluations to measure its potential for enhancing trust, transparency, and accountability in AI systems.
For this purpose, we will conduct controlled experiments, field observation and interviews, and performance evaluations~\cite{6095544}.
By gathering feedback from network experts, we aim to gain a better understanding of the strengths and weaknesses of our global explanation approach and identify areas for improvement. 
The insights gained from this evaluation will help us refine our approach to better meet the needs and expectations of domain experts and finally provide an intuitive user-friendly interactive end solution.

\section{Conclusion}
In this paper, we presented an aggregation approach for CAMs that serves as a global explanation technique for the classification of semantically structured data.
Our visualization design provides two key indicators, namely feature impact and variability, allowing for a quick overview of the global explainability of a class. 
Furthermore, the interactive approach to filter relevant samples enables a more detailed and granular level of explainability, empowering users to delve deeper into the decision-making process of the model.
We demonstrated the practical applicability of our approach by presenting a usage scenario with real data on a trained CNN model. 
Together, these advancements in aggregated CAMs offer a comprehensive and interpretable solution for the global explainability of predictions made by a CNN model, enhancing transparency and trust in its decision-making.

\section{Acknowledgments}
This research has been funded by the German Ministry of Education and Research and the Hessian State Ministry for Higher Education, Research, and the Arts as part of their support for the National Research Center for Applied Cybersecurity, ATHENE.

\clearpage
\bibliographystyle{splncs04}
\bibliography{biblio}

\end{document}